\documentclass[final,5p,times,twocolumn]{elsarticle}

\usepackage{booktabs}
\usepackage{tabularx}
\usepackage{array}
\usepackage{amsmath}
\usepackage{mathtools}
\usepackage{xcolor}
\usepackage{enumitem}
\usepackage{comment}
\usepackage{graphicx}
\usepackage{multirow}
\usepackage{array}
\usepackage{caption}
\usepackage{placeins} %
\usepackage{cuted} %
\usepackage{fontawesome5}
\usepackage{hyperref}
\usepackage{wrapfig}
\usepackage{longtable}
\usepackage{times}
\usepackage{latexsym}
\usepackage{geometry}
\usepackage{array}
\usepackage{latexsym}
\usepackage{color, colortbl}
\usepackage{booktabs}
\usepackage{multirow}
\usepackage{adjustbox}

\usepackage{booktabs}
\usepackage{multirow}
\usepackage{makecell}

\usepackage{tabularx}
\usepackage{amssymb}

\usepackage{amssymb}

\begin{document}

\begin{frontmatter}

\author[a]{Tsion Melkamu Abay\corref{cor1}}
\ead{tsioaba@mail.regent.edu}

\author[a]{Melkamu Abay Mersha\corref{cor1}}
\ead{mmersha@regent.edu}

\affiliation[a]{
    organization={Regent University},
    addressline={Virginia Beach},
    postcode={23464},
    state={VA},
    country={USA}
}

\cortext[cor1]{Corresponding authors contributed equally to this work.}

\title{Customer Segmentation and Purchase Prediction Using Explainable AI
}

\begin{abstract}
Customer segmentation can reveal differences in observed purchasing behavior without necessarily improving individual-level prediction. We investigate this distinction using a single simulated cohort of 10,000 customer records generated from fitted class-conditional Gaussian mixtures. Four classifiers are evaluated with original predictors, cluster memberships, centroid distances and matched random-anchor distances. Five stratified folds each contain 8,000 training and 2,000 test observations; nested training subsets assess sample-size sensitivity without changing the test partition. Random forest achieves mean AUROC of 0.8500 with original predictors, 0.8528 with centroid distances and 0.8530 with random-anchor distances. Its original-feature AUROC increases from 0.8264 to 0.8500 across the training-size sweep. Three-means yields the highest sampled silhouette (0.5034), while DBSCAN exhibits pronounced sensitivity to neighborhood parameters. Exact finite-background SHAP and a weighted mask-based local surrogate reveal model-specific reliance on spending and purchase history. These findings distinguish descriptive cluster quality from predictive utility and demonstrate the importance of matched nonlinear controls. Conclusions are conditional on the fitted synthetic distribution and do not establish effectiveness in independently observed customers.
\end{abstract}
\begin{keyword} Artificial Intelligence (AI), Machine Learning (ML), Explainable Artificial Intelligence (XAI), Customer Behavior Analytics, Consumer Behavior Prediction, Predictive Analytics, Customer Analytics, supervised Learning, Unsupervised Learning, Clustering Algorithms, Feature Engineering, Random Forest, LIME, SHAP  \end{keyword}
\end{frontmatter}
\nocite{piernik,ngai,chen,sakar,wedel,hennig,macqueen,lloyd,arthur,jain,mershatopic,ester,schubert,chandra,davies,calinski,rousseeuw,tibshirani,mershavirtual,hubert,hastie,quinlan,cart,breiman,rumelhart,grinsztajn,shwartz,chenxgb,gorishniy,hollmann,kohavi,varma,cawley,dietterich,bengio,nadeau,demsar,fawcett,davis,saito,brier,guocal,elkan,lipton,doshi,mershasurvey,arrieta,rudin,shapley,lundberg,ribeiro,mershaneural,sundararajan,mershacalig,covert,aas,janzing,mershaeval,adebayo,yeh,mershametrics,alvarez,slack,fisher,strobl,mershaaug,ethi,hardt,barocas,gebru,mitchell,kaufman,khapreagent,khapretox,pedregosa}

\section{Introduction}
\label{sec:introduction}

Customer purchase prediction and customer segmentation are
fundamental tasks in customer relationship management,
personalized marketing, and intelligent decision-support
systems \cite{ngai,chen,sakar}. While purchase prediction
estimates individual purchasing likelihood, segmentation
identifies groups of customers with similar characteristics
\cite{wedel,hennig}. Clustering algorithms such as K-means
\cite{macqueen,lloyd} and DBSCAN \cite{ester,schubert}
provide alternative mechanisms for discovering underlying
structures in customer data. However, the quality of the
resulting partitions depends on the feature representation,
algorithmic assumptions, and parameter configurations
\cite{jain,mershatopic}. A geometrically coherent customer
segment does not necessarily improve individual-level
prediction when the variables used to construct that
segment are already available to the classifier.

Previous research has demonstrated the potential of
clustering-based feature extraction to improve supervised
classification \cite{piernik}. Nevertheless, improvements
obtained through cluster-derived features do not necessarily
establish that learned cluster geometry is responsible for
the observed gains. Since cluster memberships and centroid
distances are transformations of existing predictors, their
predictive utility may arise from nonlinear feature expansion
rather than additional customer-level information. To
investigate this distinction, we introduce a matched-dimensionality
random-anchor control that enables a direct comparison between
distances to learned centroids and distances to randomly
selected training observations.

The effectiveness of feature augmentation may also depend
on model complexity, inductive bias, and training sample
size. Recent studies demonstrate that neural networks do
not consistently outperform conventional machine learning
models on structured tabular data
\cite{grinsztajn,shwartz,gorishniy}. Consequently, evaluating
classifiers under a single training configuration may obscure
important differences in their learning behavior. Examining
predictive performance across progressively larger training
subsets provides a more informative assessment of model
capacity and representation effectiveness.

Beyond predictive performance, understanding the factors
underlying model predictions is essential for interpreting
intelligent decision-support systems
\cite{lipton,mershasurvey,arrieta}. Although SHapley Additive
exPlanations (SHAP) \cite{lundberg} and Local Interpretable
Model-agnostic Explanations (LIME) \cite{ribeiro} are widely
used attribution methods, their explanations may differ
because of distinct mathematical formulations and
perturbation mechanisms. Recent studies emphasize the
importance of quantitatively evaluating explanation
agreement, fidelity, and sensitivity rather than relying
solely on qualitative interpretations
\cite{mershaeval,yeh,mershametrics}.

In this study, we investigate the relationships among
customer segmentation, predictive performance, training
sample size, and model explainability using a synthetic
dataset comprising 10,000 customer observations. We evaluate
Logistic Regression (LR), Decision Tree (DT), Random Forest
(RF), and Multilayer Perceptron (MLP) classifiers using
four feature representations: original predictors, cluster
memberships, centroid distances, and random-anchor distances.
We employ stratified five-fold cross-validation with
fold-local preprocessing and feature construction to
prevent predictive information leakage \cite{kaufman}.
Furthermore, we examine K-means and DBSCAN segmentation
under different parameter configurations and compare exact
finite-background SHAP with a mask-based local surrogate
(Mask-LIME).


We address the following four research questions:

\begin{itemize}

    \item \textbf{RQ1:} To what extent do cluster-derived
    feature representations improve purchase prediction
    compared with original predictors and
    matched-dimensionality random-anchor representations?

    \item \textbf{RQ2:} Which customer features receive
    the largest attributions across classification models,
    and to what extent do SHAP and Mask-LIME agree in
    identifying influential predictors?

    \item \textbf{RQ3:} How sensitive is customer segmentation
    to clustering algorithms and parameter configurations
    in terms of cluster separation, noise identification,
    and customer coverage?

    \item \textbf{RQ4:} How does training sample size
    influence predictive performance and the relative
    effectiveness of centroid-augmented representations
    under a fixed simulated distribution?

\end{itemize}


The main contributions of this study are as follows:

\begin{itemize}

    \item We develop a leakage-aware experimental framework
    for evaluating four classifiers and four feature
    representations, incorporating a random-anchor control
    to distinguish learned cluster geometry from nonlinear
    feature expansion.

    \item We characterize the influence of training sample
    size on predictive performance and cluster augmentation
    using fixed held-out partitions and consistent model
    configurations.

    \item We conduct a comparative attribution analysis
    using exact finite-background SHAP and Mask-LIME,
    examining model-specific feature reliance, explanation
    agreement, and local surrogate fidelity.

    \item We systematically evaluate K-means and DBSCAN
    configurations to distinguish descriptive segmentation
    quality from supervised predictive utility.

\end{itemize}

Our findings demonstrate that learned centroid
representations do not consistently outperform matched
random-anchor representations, despite exhibiting meaningful
descriptive clustering structures. These findings highlight
the importance of independently evaluating segmentation
quality and predictive utility. Our contribution is
empirical, and the conclusions remain conditional on the investigated synthetic distribution.


\section{Related Work}
\subsection{Customer analytics and descriptive segmentation}
Classification and clustering have long been used in customer relationship management \citep{ngai}. Chen et al. \citep{chen} characterize retail customers through recency, frequency and monetary value; Sakar et al. \citep{sakar} examine purchase intention using session-level predictors. Our dataset contains income, spending and historical activity but lacks session timestamps and a documented sampling frame, limiting direct comparisons with prospective e-commerce studies.

Segmentation depends on the variables and grouping criterion selected by the analyst \citep{wedel,hennig}. K-means minimizes within-cluster squared distances \citep{macqueen,lloyd}; its initialization and geometric assumptions influence the resulting partition \citep{arthur,jain}. Studies of transformer-based topic modeling likewise illustrate how the representation determines which similarities are available to a clustering algorithm \citep{mershatopic}. DBSCAN instead identifies density-connected regions and labels sparse observations as noise \citep{ester,schubert}. Noise is a property of the fitted density criterion, not evidence of fraud or commercial value \citep{chandra}.

Internal validation measures characterize different aspects of cohesion and separation \citep{davies,calinski,rousseeuw}. The gap statistic provides an alternative approach to cluster-count selection \citep{tibshirani}. Semantic clustering of brainstorming transcripts provides another example in which the substantive interpretation of groups depends on their representation \citep{mershavirtual}; such findings do not establish predictive utility in customer data. We therefore report partition agreement using a label-invariant measure \citep{hubert}, alongside separation and noise coverage, and evaluate cluster-derived predictors separately.

\subsection{Predictive models and evaluation design}
Model capacity and estimation error interact when training data are limited \citep{hastie}. Decision trees learn recursive partitions \citep{quinlan,cart}, random forests aggregate randomized trees \citep{breiman}, and multilayer perceptrons learn nonlinear mappings by back-propagation \citep{rumelhart}. Comparative tabular benchmarks show that deep models do not uniformly outperform tree-based methods \citep{grinsztajn,shwartz}. Gradient boosting, revised tabular neural architectures and pretrained tabular predictors offer additional alternatives \citep{chenxgb,gorishniy,hollmann}; we do not evaluate these architectures in the present study.

Piernik and Morzy \citep{piernik} investigated clustering-based feature extraction across 16 datasets. Their work motivates our use of a control representation rather than attributing any gain from appended distances to clustering itself. To separate model selection from performance estimation, nested cross-validation is often recommended; here we use fixed configurations and stratified outer folds \citep{kohavi,varma,cawley}. Shared observations induce dependence among fold-level estimates \citep{dietterich,bengio,nadeau}; multi-dataset comparisons would address a broader inferential target \citep{demsar}. We report ROC and precision--recall metrics \citep{fawcett,davis,saito} and squared probability error using the Brier score \citep{brier}. The latter does not independently establish calibration \citep{guocal}. Classification thresholds may be selected according to an explicit illustrative error-cost function \citep{elkan}.

\subsection{Explainability and attribution evaluation}
Interpretability concerns both the structure of a predictor and the information conveyed by its explanations \citep{lipton,doshi,mershasurvey,arrieta}. When transparent models achieve comparable performance, a more complex predictor requires a separate justification \citep{rudin}. SHAP allocates prediction differences using the Shapley value \citep{shapley,lundberg}; LIME approximates local behavior with a weighted surrogate \citep{ribeiro}. Neural attribution studies examine alternative mechanisms, including gradient-based relevance \citep{mershaneural,sundararajan,mershacalig}. We do not apply transformer-specific attribution to tabular models.

Removal-based explanations depend on the definition of absent features \citep{covert}. Marginal replacement can disrupt feature dependencies, while conditional and causal formulations address different estimands \citep{aas,janzing}. Quantitative studies of encoder-model explanations emphasize the need to evaluate attribution methods rather than rely solely on visual plausibility \citep{mershaeval}. Parameter sanity checks and sensitivity measures assess complementary properties \citep{adebayo,yeh}. Work on explanation agreement and evaluation metrics further motivates our comparison of rankings \citep{mershametrics,alvarez}. Because both explainers in our experiment use the same masking game, high agreement is not independent evidence of correctness. Adversarial evaluations reinforce this distinction \citep{slack}.

Permutation importance measures predictive reliance under feature perturbation \citep{fisher,strobl}. XAI-guided augmentation uses relevance to select features for controlled modification \citep{mershaaug}; retraining-based ablation can measure the effect of removing variables. Domain-specific applications, including low-resource fake-news classification \citep{ethi}, also illustrate why attribution results should be interpreted relative to the task and representation.

\subsection{Responsible decision support}
Small demographic coefficients do not establish parity in classification errors or access to opportunities \citep{hardt,barocas}. Dataset documentation and model reporting specify the population, measurement process and intended use \citep{gebru,mitchell}; these requirements are particularly important when predictors might include information unavailable at prediction time \citep{kaufman}. Broader research on trustworthy agentic systems \citep{khapreagent} and online toxicity \citep{khapretox} addresses transparency and safety in other domains. Here we examine the limits of using simulated observations for operational decisions.

\section{Data and problem definition}
We use a fitted parametric generator rather than duplicating source records. Let $Y^*\sim\mathrm{Bernoulli}(0.441)$ denote a simulated purchase label, and let $Z^*$ denote the seven standardized numerical predictors. For each class $y$, we fit a four-component Gaussian mixture to the corresponding source observations:
\begin{equation}
p(Z^*\mid Y^*=y)=\sum_{k=1}^{4}\pi_{yk}\mathcal{N}(Z^*;\mu_{yk},\Sigma_{yk}).
\end{equation}
The component covariances are regularized by adding 0.01 to their diagonals during estimation. Conditional on the sampled class and mixture component, we draw gender from a Bernoulli distribution whose probability is the responsibility-weighted proportion in that component of the source data. We invert the standardization, clip numerical values to their observed source ranges, and round count-like variables and income to integers and time and satisfaction to one decimal place. Seeds 2026 and 2027 specify generation and class-specific fitting. The generator was calibrated on a privately supplied educational dataset of 1,000 records; these are not an evaluation cohort. Its collection and label-generation processes are undocumented. The resulting file contains 10,000 unique identifiers, 4,482 positive and 5,518 negative labels, no missing values and no duplicate feature-label rows. Labels are generated first; the simulation therefore reproduces a fitted statistical association rather than a temporal or causal purchase process. It is not a privacy guarantee or an independent customer sample.

Let $D=\{(x_i,y_i)\}_{i=1}^{10000}$, where $x_i\in\mathbb{R}^{8}$ and $y_i\in\{0,1\}$. Identifiers are excluded; gender is encoded as 0 (Female) and 1 (Male). The task is to rank the simulated binary outcome on held-out records. Annual income and spending score define the descriptive clustering space. No purchase labels enter the clustering fit. Because labels are sampled before features, the experiment does not establish temporal or causal prediction validity.
\begin{table}[t]\centering\small\caption{Numerical characteristics of the sole experimental cohort ($N=10,000$).}\label{tab:data}
\begin{adjustbox}{max width=\columnwidth}\begin{tabular}{lrrr}\toprule Feature & Range & Mean & SD\\\midrule
Age & 18.0--73.0 & 39.91 & 10.89\\
Annual Income & 20,000.0--137,673.0 & 68,100.17 & 28,948.70\\
Spending Score & 1.0--100.0 & 50.77 & 23.13\\
Website Visits & 1.0--20.0 & 7.58 & 3.40\\
Time On Website & 1.0--17.9 & 7.56 & 2.96\\
Previous Purchases & 0.0--17.0 & 4.75 & 3.32\\
Satisfaction Score & 1.7--10.0 & 6.92 & 1.50\\
\bottomrule\end{tabular}\end{adjustbox}\end{table}
The cohort contains 5,198 Female and 4,802 Male records. Table~\ref{tab:data} summarizes the numerical variables.
\section{Methodology}
\begin{figure*}[t]\centering\includegraphics[width=.75\linewidth]{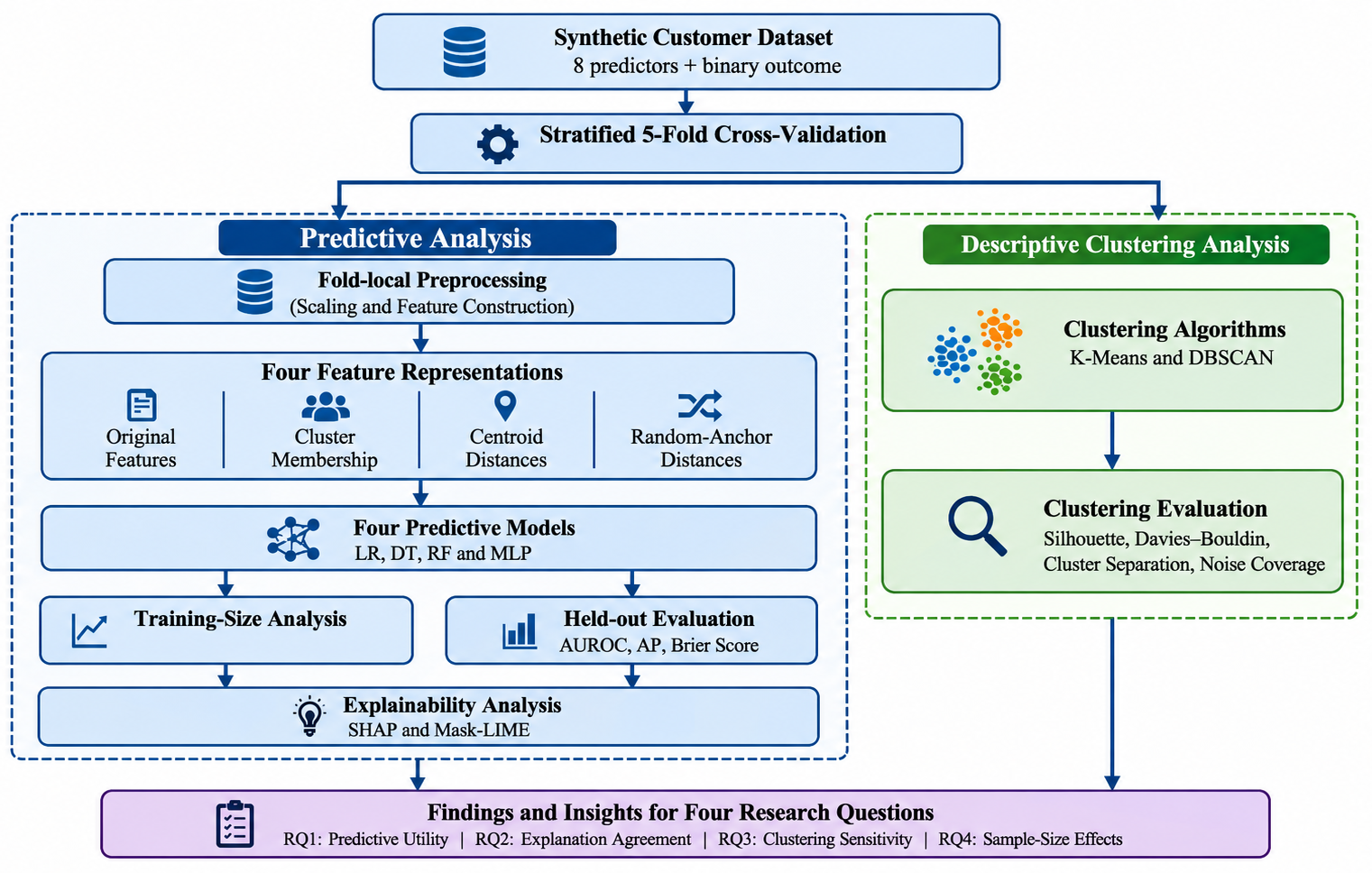}\caption{Experimental framework for the single simulated cohort. All predictive transformations are fitted within the selected training subset.}\label{fig:framework}\end{figure*}

Figure~\ref{fig:framework} separates descriptive clustering from the held-out predictive and attribution branches.
\subsection{Fold-local feature representations}
For a training partition $T$, each predictor is standardized using its training mean and standard deviation, $z_j=(x_j-\mu_{T,j})/\sigma_{T,j}$. Let $u=(z_{\mathrm{income}},z_{\mathrm{spending}})$. We estimate four centroids $c_1,\ldots,c_4$ using K-means with ten initializations and seed 42 by minimizing
\begin{equation}
 \sum_{i\in T}\min_{1\leq j\leq 4}\|u_i-c_j\|_2^2.
\end{equation}
Four representations are evaluated: original standardized predictors $z$; hard assignment $[z;e_{a(u)}]$, where $a(u)=\arg\min_j\|u-c_j\|_2$ and $e_j$ is a four-dimensional one-hot vector; centroid distances $[z;d(u)]$ with $d_j(u)=\|u-c_j\|_2$; and random-anchor distances $[z;r(u)]$ using four distinct training rows sampled without replacement with seed 42. The distance representations have the same dimensionality, but they differ in how anchors are chosen. All new features are deterministic functions of the original predictors and fitted training statistics. Therefore, conditional on a fitted transformation, they cannot add independent customer-level information; their utility can arise through changes in model inductive bias, regularization or finite-sample approximation.
\subsection{Classifiers and evaluation}
We use fixed configurations throughout the training-size sweep to isolate sample availability from changes in tuning. Logistic regression uses $C=1$ and at most 1000 iterations; the decision tree uses depth 5 and minimum leaf size 10; random forest uses 30 trees, depth 8 and minimum leaf size 5. The MLP uses ReLU hidden layers (16,8), Adam, ridge penalty 0.01, at most 100 iterations and internal early stopping with a 0.1 validation fraction. Its effective gradient-training sample is therefore smaller than the nominal subset. Preprocessing precedes this internal validation split, but outer test observations remain excluded. All models use the same five stratified folds (seed 42), with 8,000 training and 2,000 held-out observations per fold. Nested stratified training subsets contain 500, 1,000, 2,000, 4,000 and 8,000 observations. Raw and centroid-distance features are evaluated at every size; membership and random anchors are additionally evaluated at the largest size. This yields 240 model--fold fits. We report AUROC, average precision, accuracy, positive-class F1, recall and Brier score; thresholded metrics use 0.5. Fold standard deviations describe variability rather than independent-sample confidence intervals. The comparison is exploratory and does not establish equivalence.
\subsection{Descriptive clustering}
We standardize income and spending on the full simulated cohort for the descriptive analysis only. K-means uses $K=2,\ldots,8$; DBSCAN uses $\epsilon\in\{.10,.15,.20,.25,.30\}$ and minimum samples in $\{8,20,50,80\}$. We report cluster counts, noise, Davies--Bouldin indices and silhouette estimates computed on at most 2,000 observations. DBSCAN indices exclude noise and are undefined for a single cluster. These descriptive fits never supply predictive parameters.
\subsection{Model-agnostic local explanations}
We use exact finite-background interventional SHAP \citep{lundberg} and a mask-based implementation of the local surrogate principle of LIME \citep{ribeiro}. For each outer fold, five held-out records and 32 training background records are sampled without replacement using seed $300+i$ (after partition shuffling). The same records and background are used across four raw-feature classifiers. For each explained record $x$ and subset $S\subseteq N=\{1,\ldots,8\}$, define
\begin{equation}
v_x(S)=\frac{1}{32}\sum_{b\in B} f(x_S,b_{N\setminus S}).
\end{equation}
The model output is the positive-class probability. We enumerate all 256 subsets and calculate
\begin{equation}
\begin{aligned}
\phi_j(x)&=\sum_{S\subseteq N\setminus\{j\}}\frac{|S|!(7-|S|)!}{8!}\\
&\quad\times[v_x(S\cup\{j\})-v_x(S)].
\end{aligned}
\end{equation}
This is exact for the empirical background game, not for an unknown population distribution. The masking can break feature dependencies and generate unrealistic hybrid customers. For the local surrogate (Mask-LIME), let $m_S$ be the binary indicator of retained features and use weighted ridge regression:
\begin{equation}
\begin{aligned}
\min_{\beta_0,\beta}\;&\sum_S e^{-(8-|S|)/2}\\
&\times(v_x(S)-\beta_0-m_S^\top\beta)^2\\
&+0.01\|\beta\|_2^2.
\end{aligned}
\end{equation}
The intercept is unpenalized. This transparent variant uses exhaustive binary masks and background averaging instead of the default tabular LIME package's sampling and discretization; its coefficients should not be represented as package-default LIME outputs. We report mean absolute attributions across 25 fold-record instances per classifier, weighted surrogate $R^2$ on the fitted coalition set, absolute approximation error at the original record, SHAP reconstruction error, Spearman correlation of absolute attribution rankings, and top-three feature overlap. Shared masks make the comparison controlled but may favor agreement. The diagnostic is not independent perturbation validation.

\section{Results}
\subsection{Predictive performance and learning curves}

\begin{table*}[t]\centering\small\caption{Simulated cohort: AUROC (mean $\pm$ fold SD) at 8,000 training and 2,000 test records per fold. Five stratified folds; fixed model settings. Membership adds four one-hot coordinates, whereas both distance representations add four continuous coordinates. Higher AUROC indicates better ranking; fold SD describes partition variability.}\label{tab:scale-full}
\begin{tabular}{lrrrr}\toprule Model & Original & Membership & Centroid distances & Random distances\\\midrule
LR & $0.8349 \pm 0.0108$ & $0.8356 \pm 0.0107$ & $0.8359 \pm 0.0103$ & $0.8360 \pm 0.0110$\\
Tree & $0.8221 \pm 0.0129$ & $0.8225 \pm 0.0125$ & $0.8227 \pm 0.0120$ & $0.8225 \pm 0.0091$\\
RF & $0.8500 \pm 0.0112$ & $0.8525 \pm 0.0129$ & $0.8528 \pm 0.0113$ & $0.8530 \pm 0.0103$\\
MLP & $0.8284 \pm 0.0162$ & $0.8173 \pm 0.0215$ & $0.8363 \pm 0.0098$ & $0.8371 \pm 0.0102$\\
\bottomrule\end{tabular}\end{table*}

\begin{figure}[t]\centering\includegraphics[width=0.47\textwidth]{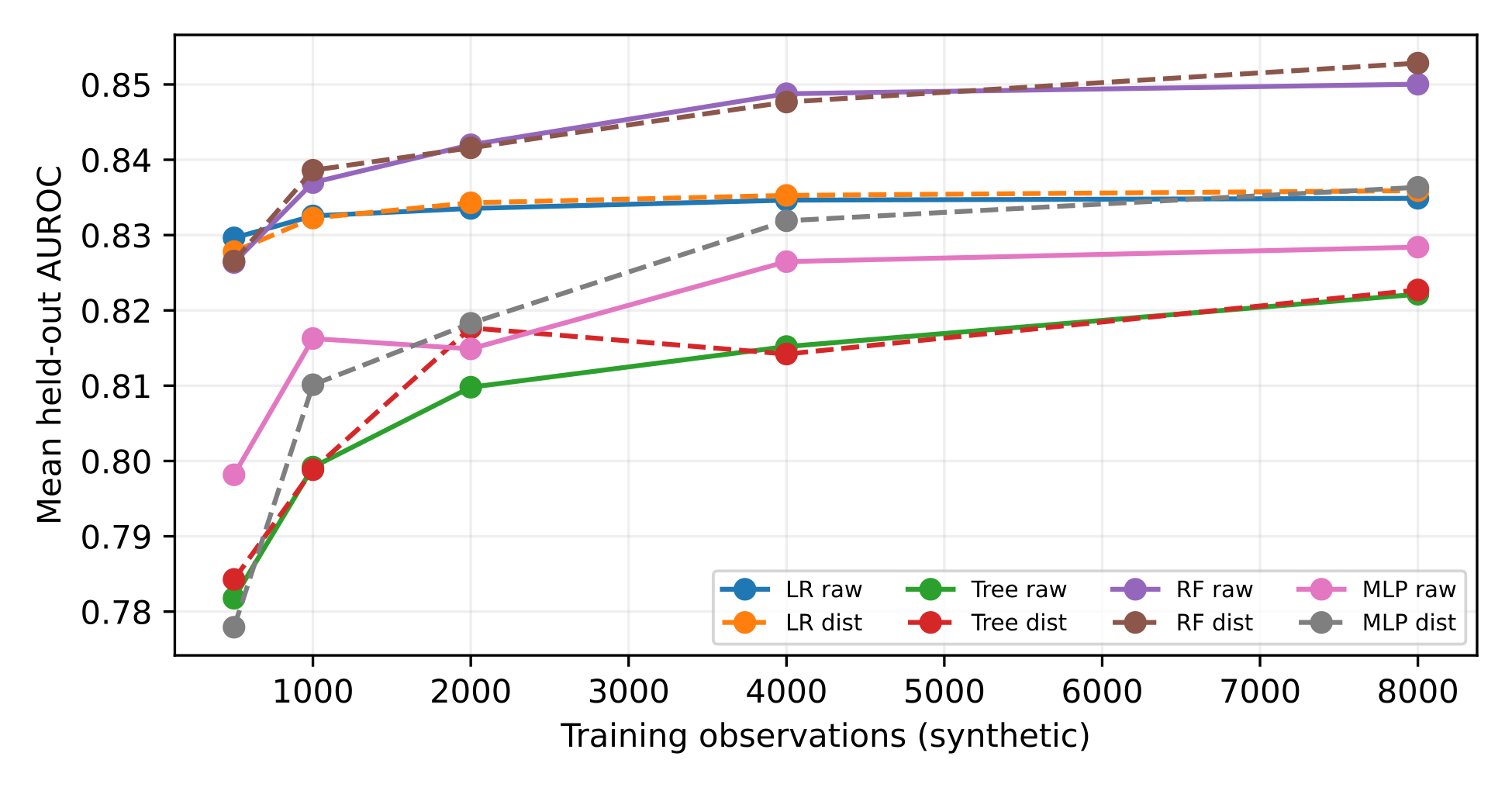}\caption{Learning curves within the synthetic cohort. Solid lines use original predictors and dashed lines add distances to learned centroids. Each point averages five folds; test partitions are fixed within folds across training sizes. The curves compare sample availability under fixed model settings; random-anchor and membership controls are evaluated only at 8,000 records.}\label{fig:scale}
\end{figure}
\begin{figure*}[t]\centering\includegraphics[width=.6\linewidth]{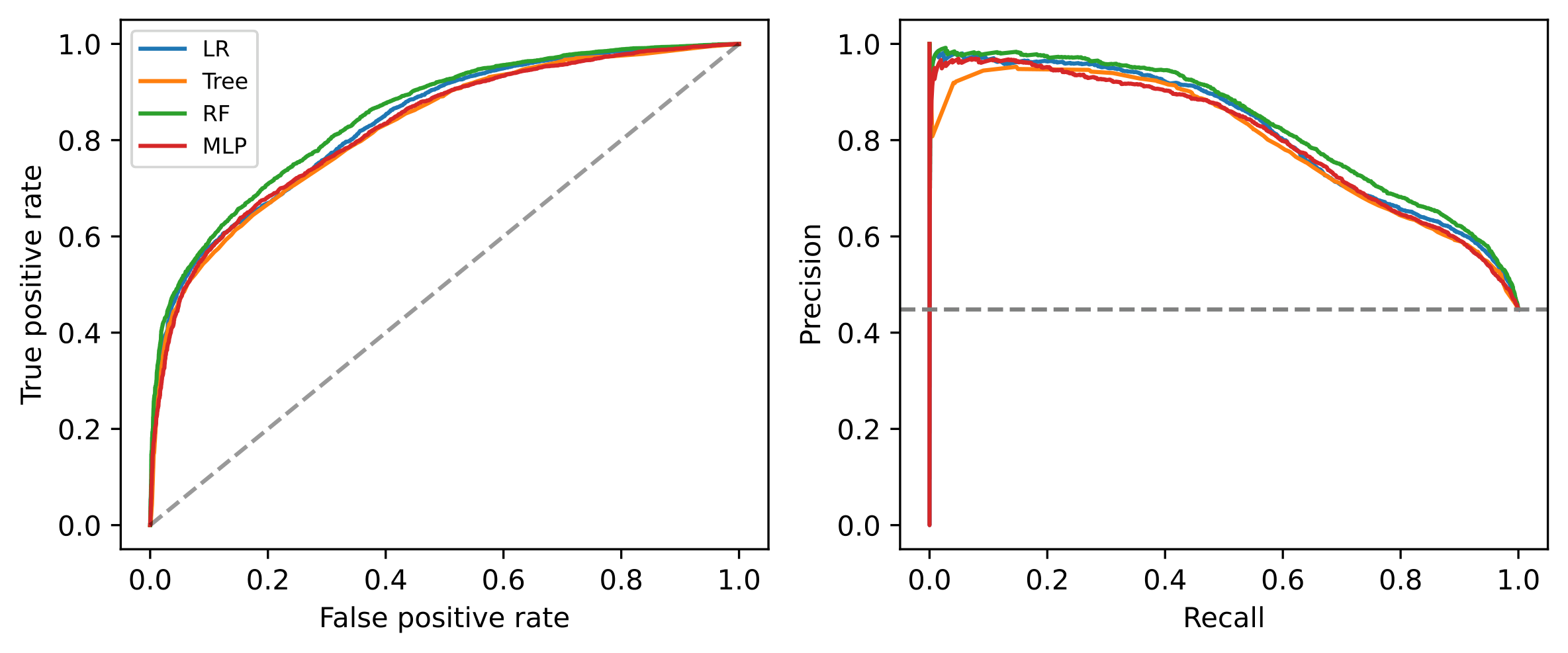}\caption{Pooled out-of-fold ROC and precision--recall curves at the largest training size, using original predictors. Left: ROC with the random-ranking diagonal. Right: precision--recall with the positive-prevalence reference (0.4482).}\label{fig:roc}\end{figure*}
Figure~\ref{fig:roc} illustrates ranking performance across thresholds; pooled curves complement, rather than replace, the fold-level means. The ROC curves lie above the diagonal reference, indicating discrimination beyond random ranking. The RF curve generally occupies the upper region of the plot, consistent with its higher fold-mean AUROC. In the precision--recall panel, the horizontal reference corresponds to the positive prevalence of 0.4482. Precision generally declines as recall approaches one, illustrating the trade-off between identifying more simulated purchasers and retaining a high proportion of correct positive predictions. The raw-feature RF achieves average precision (AP) of 0.8382, compared with 0.8214 for LR, 0.8115 for MLP and 0.7917 for DT (Table~\ref{tab:full-metrics}). These results support the ranking advantage of RF within the experiment, but the pooled curves do not quantify uncertainty at individual thresholds or establish a deployment threshold.

Table~\ref{tab:scale-full} shows that RF attains the highest mean AUROC in the simulated cohort. With original predictors, RF improves by 0.0236 between 500 and 8,000 training observations, while LR improves by only 0.0053. DT and MLP improve by 0.0403 and 0.0302, respectively, over the same range. These are descriptive differences from Figure~\ref{fig:scale}, not evidence of a universal sample-size effect. MLP centroid augmentation adds 0.0079 at the largest size, but random-anchor distances add 0.0087. RF exhibits a similar pattern: 0.0028 versus 0.0030. Thus even in the larger simulation, a useful nonlinear expansion does not uniquely establish a contribution from meaningful segments. All learning curves are conditional on the same fitted generator and fixed model settings.

\paragraph{Effect of feature representation.} At the largest training size, the centroid-distance gains over raw features are 0.0010, 0.0006, 0.0028 and 0.0079 for LR, DT, RF and MLP, respectively. Random-anchor distances give corresponding gains of 0.0011, 0.0004, 0.0030 and 0.0087. For three of the four classifiers, the random representation attains a slightly higher mean AUROC than the learned-centroid representation; for DT the ordering is reversed by only 0.0002. The hard-membership representation yields modest gains for LR, DT and RF but decreases MLP AUROC from 0.8284 to 0.8173. Unlike continuous distances, membership discards within-region variation in the added representation and changes abruptly at a cluster boundary. This is a plausible explanation for the different behavior, not a mechanism directly tested by the experiment. Because both distance representations retain the original predictors and add four derived coordinates, the comparison controls for feature count but not every aspect of feature geometry or optimization.

\begin{table*}[t]
\scriptsize
\centering\small\caption{Full-training AUROC per fold and mean difference from raw features.}\label{tab:fold-variation}
\begin{tabular}{llrrrrrr}\toprule Model & Mode & Fold 1 & Fold 2 & Fold 3 & Fold 4 & Fold 5 & $\Delta$ AUROC\\\midrule
LR & dist & 0.8320 & 0.8199 & 0.8460 & 0.8418 & 0.8398 & +0.0010\\
LR & hard & 0.8315 & 0.8190 & 0.8465 & 0.8415 & 0.8393 & +0.0007\\
LR & random & 0.8312 & 0.8194 & 0.8479 & 0.8406 & 0.8409 & +0.0011\\
LR & raw & 0.8307 & 0.8185 & 0.8467 & 0.8405 & 0.8381 & +0.0000\\
MLP & dist & 0.8424 & 0.8198 & 0.8346 & 0.8431 & 0.8418 & +0.0079\\
MLP & hard & 0.8090 & 0.7851 & 0.8192 & 0.8373 & 0.8357 & -0.0111\\
MLP & random & 0.8369 & 0.8210 & 0.8381 & 0.8492 & 0.8403 & +0.0087\\
MLP & raw & 0.8113 & 0.8200 & 0.8489 & 0.8197 & 0.8421 & +0.0000\\
RF & dist & 0.8544 & 0.8371 & 0.8677 & 0.8571 & 0.8479 & +0.0028\\
RF & hard & 0.8506 & 0.8331 & 0.8670 & 0.8608 & 0.8510 & +0.0025\\
RF & random & 0.8545 & 0.8371 & 0.8651 & 0.8572 & 0.8512 & +0.0030\\
RF & raw & 0.8495 & 0.8319 & 0.8614 & 0.8565 & 0.8510 & +0.0000\\
Tree & dist & 0.8235 & 0.8087 & 0.8395 & 0.8277 & 0.8141 & +0.0006\\
Tree & hard & 0.8189 & 0.8078 & 0.8392 & 0.8312 & 0.8157 & +0.0004\\
Tree & random & 0.8212 & 0.8124 & 0.8355 & 0.8272 & 0.8164 & +0.0004\\
Tree & raw & 0.8164 & 0.8085 & 0.8404 & 0.8302 & 0.8152 & +0.0000\\
\bottomrule\end{tabular}
\end{table*}

Table~\ref{tab:fold-variation} reports the fold-level results. The fold-level differences are paired because each representation is evaluated on the same test partition. The five folds share training observations, so their variability is not a formal independent-sample significance test. The fold-level results qualify the mean improvements. RF centroid distances improve AUROC in four of five folds, whereas random distances improve it in all five using the reported rounded values. For MLP, centroid distances improve two folds and decrease three, despite a positive mean gain. In particular, the increase from 0.8113 to 0.8424 in the first fold contrasts with the decrease from 0.8489 to 0.8346 in the third. The MLP fold SD also decreases from 0.0162 to 0.0098 with centroid distances, but this is descriptive variation across partitions, not evidence of robustness to new populations or training seeds. The small differences between the two distance representations should not be interpreted as statistically established advantages.

\paragraph{Training-size dependence.} Figure~\ref{fig:scale} reveals distinct learning profiles. LR starts at 0.8296 with 500 records and reaches 0.8349 with 8,000, suggesting that much of its attainable discrimination under the fixed specification is already captured by the smaller subset. RF overtakes LR by 1,000 records and reaches 0.8500 at the largest size. DT gains the most in raw-feature AUROC over the full sweep, although its final value remains below RF. The augmented MLP initially performs worse than its raw counterpart (0.7779 versus 0.7982 at 500 records), but exceeds it at 2,000, 4,000 and 8,000 records. The difference changes from $-0.0203$ to $+0.0079$ across the sweep. This pattern is compatible with additional derived features requiring more observations for useful estimation under the fixed architecture; optimization and internal early stopping may also contribute. Neither explanation is isolated experimentally. The nonmonotonic segments of the curves and the near-plateau of LR caution against extrapolation beyond 8,000 records. Random anchors and hard membership were not evaluated across the smaller sizes, so a sample-size-specific advantage of learned centroids over these controls cannot be established.

\begin{table*}[t]
\scriptsize
\centering\small\caption{All synthetic full-training classification metrics (five-fold means).}\label{tab:full-metrics}
\begin{tabular}{llrrrrrr}\toprule Model & Features & AUROC & AP & Accuracy & F1 & Recall & Brier\\\midrule
LR & raw & 0.8349 & 0.8214 & 0.7472 & 0.6950 & 0.6428 & 0.1629\\
LR & hard & 0.8356 & 0.8220 & 0.7524 & 0.6959 & 0.6323 & 0.1623\\
LR & dist & 0.8359 & 0.8202 & 0.7545 & 0.6994 & 0.6374 & 0.1620\\
LR & random & 0.8360 & 0.8210 & 0.7488 & 0.6961 & 0.6419 & 0.1623\\
Tree & raw & 0.8221 & 0.7917 & 0.7468 & 0.6831 & 0.6109 & 0.1667\\
Tree & hard & 0.8225 & 0.7928 & 0.7470 & 0.6831 & 0.6102 & 0.1666\\
Tree & dist & 0.8227 & 0.7887 & 0.7465 & 0.6622 & 0.5544 & 0.1663\\
Tree & random & 0.8225 & 0.7912 & 0.7470 & 0.6712 & 0.5772 & 0.1664\\
RF & raw & 0.8500 & 0.8382 & 0.7632 & 0.6998 & 0.6167 & 0.1548\\
RF & hard & 0.8525 & 0.8410 & 0.7647 & 0.7008 & 0.6158 & 0.1540\\
RF & dist & 0.8528 & 0.8408 & 0.7682 & 0.7056 & 0.6209 & 0.1540\\
RF & random & 0.8530 & 0.8404 & 0.7648 & 0.7029 & 0.6212 & 0.1537\\
MLP & raw & 0.8284 & 0.8115 & 0.7518 & 0.6929 & 0.6249 & 0.1676\\
MLP & hard & 0.8173 & 0.8062 & 0.7538 & 0.6864 & 0.6017 & 0.1681\\
MLP & dist & 0.8363 & 0.8206 & 0.7508 & 0.6905 & 0.6214 & 0.1624\\
MLP & random & 0.8371 & 0.8227 & 0.7536 & 0.6967 & 0.6327 & 0.1617\\
\bottomrule\end{tabular}
\end{table*}

\paragraph{Complementary predictive metrics.} Table~\ref{tab:full-metrics} shows that AUROC improvements do not necessarily translate into improved decisions at the fixed threshold of 0.5. RF with centroid distances achieves the highest reported accuracy (0.7682) and F1 (0.7056), whereas RF with random distances obtains the highest AUROC (0.8530) and lowest Brier score (0.1537). RF membership has the highest AP (0.8410). LR with raw features has the highest recall among the reported configurations (0.6428), despite lower AUROC than RF. For DT, centroid distances increase AUROC slightly but reduce recall from 0.6109 to 0.5544 and F1 from 0.6831 to 0.6622. MLP centroid distances improve AUROC and Brier score (0.1676 to 0.1624), but reduce F1 slightly (0.6929 to 0.6905). These contrasts show why discrimination, threshold-dependent classification and probability quality should be assessed separately. A lower Brier score indicates lower mean squared probability error, but does not by itself demonstrate calibration across probability ranges. No single configuration dominates every metric.

\subsection{Segmentation and explanation diagnostics}

Three-means achieves a sampled silhouette of 0.5034; four-means achieves 0.4965. At $\epsilon=0.20$ and minimum samples 8, DBSCAN yields one cluster and 32 noise observations, so its silhouette is undefined. Smaller neighborhoods or higher density requirements fragment the partition and can substantially reduce coverage. Table~\ref{tab:cluster-sensitivity} provides the entire sweep. Four centroids remain fixed in the predictive comparison to match the four random anchors; we do not select the predictive representation on the basis of the full-cohort silhouette.

\begin{figure*}[t]\centering\includegraphics[width=.75\linewidth]{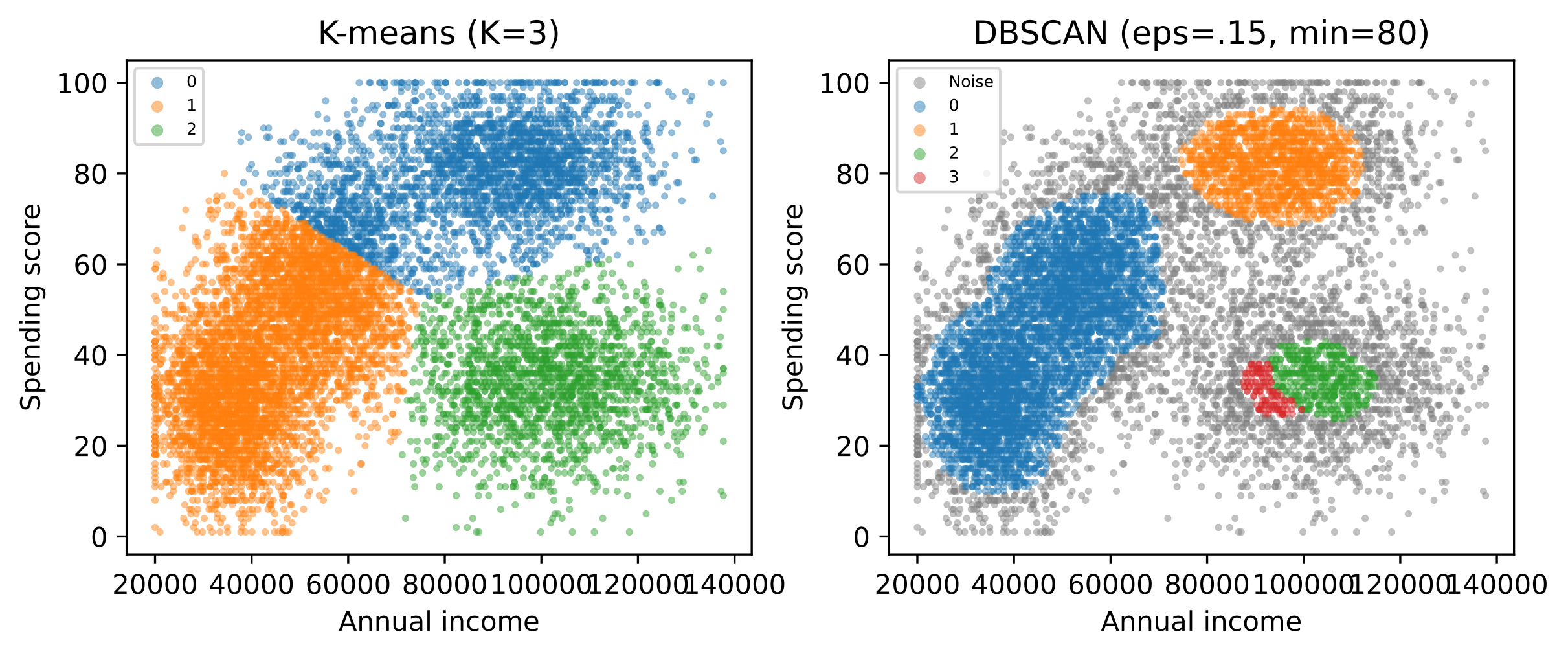}\caption{Descriptive segmentation of the 10,000 simulated records in income--spending space. Left: three-means with complete assignment. Right: DBSCAN ($\epsilon=0.15$, minimum samples 80), with four density-based groups and 3,217 noise observations (gray).}\label{fig:segments}\end{figure*}

Figure~\ref{fig:segments} compares three-means with DBSCAN at $\epsilon=0.15$ and minimum samples 80. The K-means panel partitions the entire income--spending plane into regions broadly characterized by lower income, higher spending, and higher income with lower spending. The boundaries reflect nearest-centroid assignments rather than demonstrated discontinuities in customer behavior. The DBSCAN panel instead identifies dense cores and leaves dispersed observations unassigned. Its four clusters cover 6,783 records, while 3,217 (32.17\%) are noise. The silhouette of 0.4820 applies only to assigned observations and is therefore not directly comparable to the full-coverage K-means value of 0.5034. The higher-density requirement can yield visually compact groups at the cost of excluding substantial portions of the cohort.

The sensitivity sweep further shows that the three-means solution has the highest reported K-means silhouette and lowest Davies--Bouldin index (0.6573), with four-means close behind (0.4965 and 0.6597). Increasing $K$ to eight reduces silhouette to 0.3461 and raises Davies--Bouldin to 1.0170, suggesting that additional partitions do not necessarily improve internal separation. At $\epsilon=0.10$ and minimum samples 80, DBSCAN assigns only 1,970 observations while attaining a silhouette of 0.4651. At $\epsilon=0.25$ and 0.30, all tested minimum-sample settings produce a single cluster. Together these results indicate that the apparent number and quality of density-based groups are highly conditional on the parameter choice. Neither internal metric measures purchase-label separation, temporal segment stability or business usefulness.
We compute exact finite-background SHAP and Mask-LIME for the raw classifiers at 8,000 training records, using five held-out observations per fold and a 32-row training background. The 100 model-instance explanations use all 256 masks. Table~\ref{tab:scale-xai} reports the mean agreement and fidelity measures; the methods share a masking game, so these diagnostics do not constitute independent validation of explanation truth. Previous purchases has the largest mean absolute SHAP magnitude for LR and MLP, while spending score ranks first for DT and RF. These results characterize models fitted to the generator, not causal customer behavior. We do not claim a size-dependent explanation stability effect because attributions were evaluated only at the largest training size.
\begin{table}[t]\centering\small\caption{Synthetic-cohort explanation diagnostics (25 cases per model). $\rho$ denotes Spearman agreement of absolute attribution rankings, Top-3 the mean top-three overlap, $R^2_w$ weighted surrogate fit, and point error absolute approximation error at the original record (probability units).}\label{tab:scale-xai}\begin{adjustbox}{max width=\columnwidth}\begin{tabular}{lrrrr}\toprule Model & $\rho$ & Top-3 & $R^2_w$ & Point error\\\midrule
LR & 0.9895 & 0.9733 & 0.9932 & 0.0090\\
Tree & 0.9743 & 0.9467 & 0.8501 & 0.0216\\
RF & 0.9676 & 0.9467 & 0.9407 & 0.0178\\
MLP & 0.9667 & 0.9867 & 0.9413 & 0.0127\\
\bottomrule\end{tabular}\end{adjustbox}\end{table}

\begin{figure*}[t]\centering\includegraphics[width=.75\linewidth]{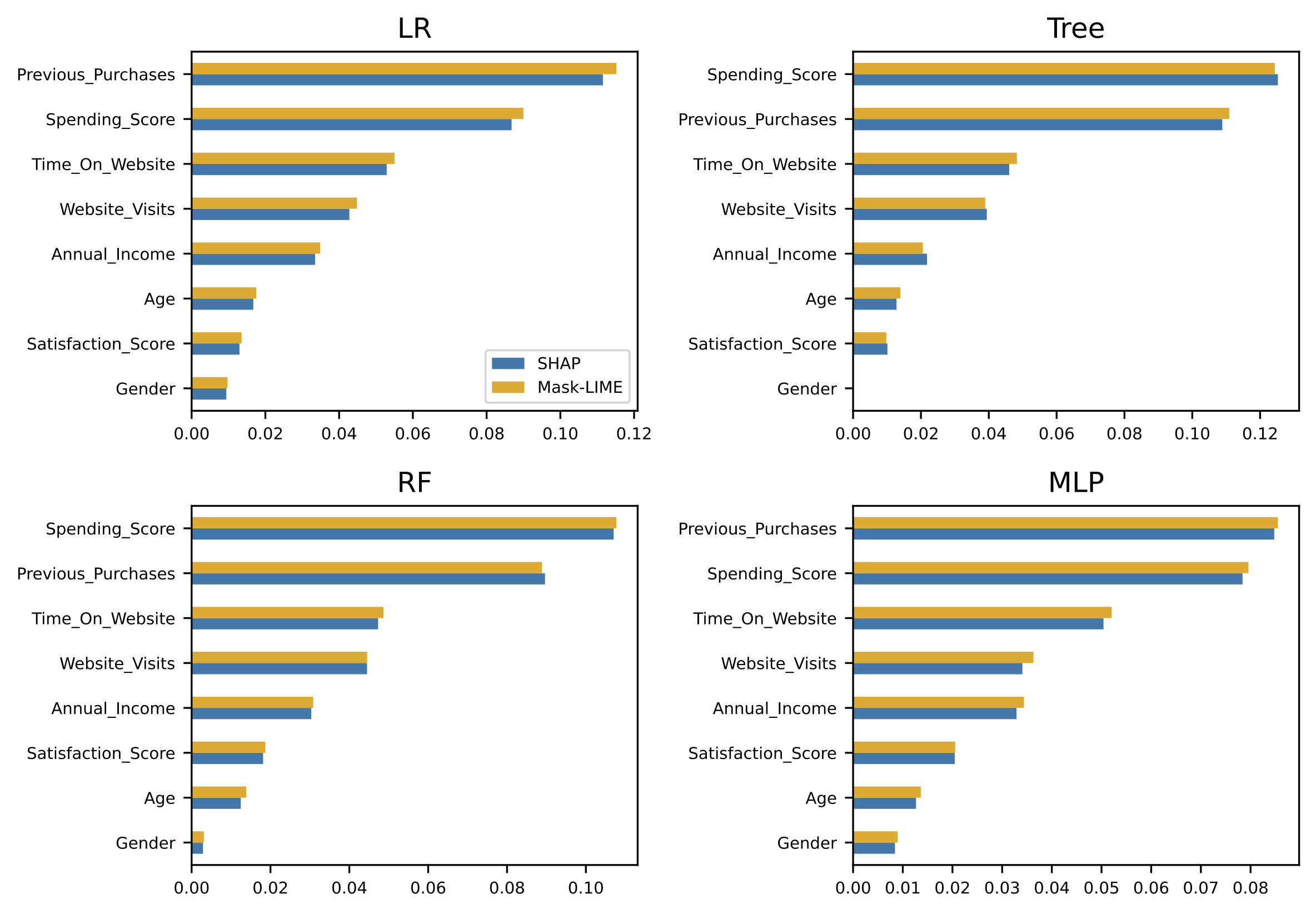}\caption{Mean absolute SHAP and Mask-LIME attributions for the four raw-feature classifiers, 25 held-out cases each. Bars represent attribution magnitudes in probability units, not signed effects or causal importance. Panel scales may differ.}\label{fig:global}\end{figure*}
\begin{figure*}[t]\centering\includegraphics[width=.75\linewidth]{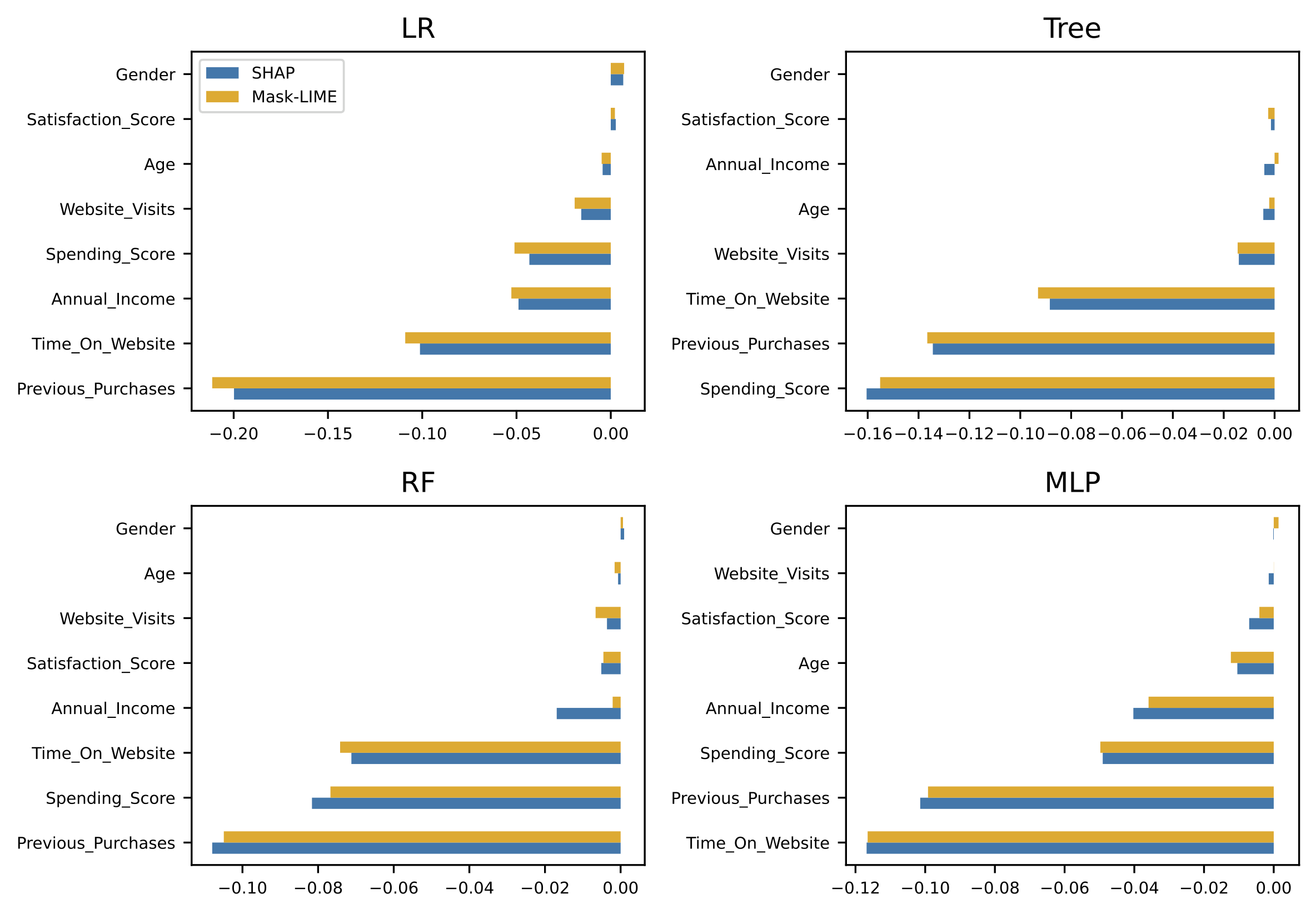}\caption{Example signed attributions for a shared held-out case. Negative values indicate contributions below the model-specific background reference in the empirical masking game; positive values indicate the opposite. Panel scales differ. These values describe the fitted models, not causal effects.}\label{fig:local}\end{figure*}

\paragraph{Agreement and surrogate fidelity.} Table~\ref{tab:scale-xai} distinguishes ranking agreement from the ability of the fitted surrogate to reproduce masked predictions. Spearman correlations range from 0.9667 to 0.9895, and mean top-three overlap ranges from 0.9467 to 0.9867. LR has the strongest rank correlation and weighted surrogate fit ($R^2_w=0.9932$), with a mean original-point error of 0.0090 in probability units. DT has a comparatively lower weighted fit (0.8501) and the largest point error (0.0216), although its rank correlation remains high (0.9743). The RF and MLP weighted fits are similar (0.9407 and 0.9413), but MLP has a smaller point error (0.0127 versus 0.0178). The MLP also has the highest top-three overlap despite having the lowest overall rank correlation. Thus agreement on the leading features can coexist with differences elsewhere in the ranking. The tree's piecewise-constant decision surface may be harder to approximate with an additive surrogate across masks, but the experiment does not isolate interaction or discontinuity effects. These diagnostics concern the shared empirical coalition game, not independent explanation fidelity under realistic customer perturbations.

\paragraph{Aggregate attribution patterns.} Figure~\ref{fig:global} displays mean absolute attributions in probability units. Previous purchases and spending score are the two leading features for all four classifiers, and time on website ranks third. ~\ref{app:xai} gives the corresponding values: for LR, previous purchases contributes a mean absolute SHAP magnitude of 0.1115 compared with 0.0868 for spending score; for DT, spending score reaches 0.1252 compared with 0.1089 for previous purchases. RF similarly places spending score first (0.1070 versus 0.0896), whereas MLP places previous purchases first (0.0848 versus 0.0784). The broad agreement suggests that these fitted classifiers rely on a similar subset of variables, but relative emphasis varies with the model. Income ranks below website engagement measures in the aggregate, despite being one of the two clustering coordinates. This helps distinguish the geometry used to construct segments from the features most influential for the raw classifiers. Gender has the smallest mean absolute SHAP magnitude across models, including zero for DT at the reported precision; this does not establish fairness or absence of indirect demographic effects. Absolute values remove direction, so the global panels cannot indicate whether a feature increases or decreases a particular prediction. Moreover, these summaries aggregate only 25 held-out instances per classifier, not the full cohort.

\paragraph{Individual prediction interpretation.} Figure~\ref{fig:local} illustrates how signed contributions differ for a common held-out case. Previous purchases is the largest negative contribution for LR and RF, whereas spending score is the largest negative contribution for DT and time on website is the largest for MLP. For LR, the previous-purchases SHAP contribution is approximately $-0.20$ probability units, while the corresponding RF contribution is approximately $-0.11$. These are contributions relative to each model's empirical background prediction, not changes caused by intervening on the customer. The mostly negative bars indicate that the retained feature values tend to lower the modeled probability relative to the corresponding background game for this case. The two methods agree on the dominant negative factors in each panel, although smaller contributions can differ in magnitude or sign. For example, the RF income contribution is visibly more negative under SHAP than under Mask-LIME. Such local differences are obscured by aggregate absolute magnitudes. The MLP example also demonstrates that the globally highest-ranked feature need not dominate every prediction. Since panel scales and model baselines differ, the figure should not be read as a comparison of absolute predicted probabilities. No case-level probability or customer outcome is inferred from the bars alone. Finally, the explanations concern raw-feature models; they cannot establish whether the added centroid distances are used by, or faithfully explain, the augmented classifiers.

\section{Discussion}

\paragraph{RQ1: Predictive effectiveness of customer segmentation.}
Our results demonstrate that the predictive benefits of cluster-derived
features vary across classifiers and feature representations. RF achieves
the highest mean AUROC, increasing from 0.8500 with original predictors
to 0.8528 with learned-centroid distances and 0.8530 with random anchors.
MLP exhibits larger improvements of 0.0079 and 0.0087, respectively,
whereas hard membership reduces its AUROC by 0.0111. LR and DT show
only marginal improvements. Notably, random-anchor distances achieve
higher mean AUROC than learned-centroid distances in three of the four
classifiers. The fold-level results further reveal that these gains
are not consistent across all partitions.

These findings suggest that predictive improvements may arise from
nonlinear feature transformation rather than learned customer segment
structure. Furthermore, RF with random anchors achieves the highest
AUROC and lowest Brier score, while centroid distances yield the
highest accuracy and F1 score. For DT, centroid augmentation slightly
improves AUROC but reduces recall from 0.6109 to 0.5544. Thus, improved
ranking performance does not necessarily translate into better
classification decisions, and our results do not establish a unique
predictive advantage for learned customer segments.

\paragraph{RQ2: Feature importance and explanation reliability.}
Our explainability analysis consistently identifies previous purchases,
spending score, and time on website as the three leading predictors.
Previous purchases dominates LR and MLP, whereas spending score
dominates DT and RF. Income, despite being a segmentation variable,
ranks below these behavioral attributes, suggesting that variables
defining customer segments are not necessarily the strongest
predictive features.

Although SHAP and Mask-LIME exhibit strong rank correlations
(0.9667--0.9895), their surrogate fidelity varies considerably.
For example, DT achieves a lower weighted surrogate fit (0.8501)
than LR (0.9932). Our local explanations also reveal differences
in feature contributions across classifiers. These findings
highlight the distinction between attribution agreement and
explanation fidelity. However, their shared masking mechanism
limits the independence of this comparison, and neither method
establishes causal feature importance.

\paragraph{RQ3: Effectiveness of customer segmentation.}
Our clustering experiments demonstrate that segmentation quality
depends on algorithm selection, parameter configuration, and
population coverage. Three-means achieves the highest silhouette
score (0.5034) and lowest Davies--Bouldin index (0.6573) among the
evaluated K-means configurations. Conversely, DBSCAN achieves a
silhouette score of 0.4820 while excluding 32.17\% of customers
as noise. Another configuration excludes 80.30\% while maintaining
a silhouette score of 0.4651. These results demonstrate that
favorable internal clustering scores can conceal limited population
coverage. Moreover, the strongest descriptive partition does not
necessarily provide the optimal predictive representation.
We therefore emphasize evaluating descriptive cluster quality
and predictive usefulness as distinct objectives.

\paragraph{RQ4: Influence of training sample size.}
Increasing the training subset from 500 to 8,000 records improves
the original-feature AUROC of DT, MLP, RF, and LR by 0.0403,
0.0302, 0.0236, and 0.0053, respectively. These results indicate
that LR approaches a performance plateau earlier than the other
classifiers. Interestingly, centroid augmentation reduces MLP
AUROC by 0.0203 at 500 records but improves it by 0.0079 at
8,000 records, suggesting that the effectiveness of distance-based
augmentation may depend on training sample availability. However,
the absence of random-anchor comparisons at smaller training
sizes prevents attributing this behavior specifically to learned
centroids.

\paragraph{Overall implications.}
Our findings emphasize the importance of matched nonlinear controls
when assessing the predictive contribution of customer segmentation.
They further demonstrate that descriptive cluster quality,
predictive effectiveness, and explanation reliability represent
distinct evaluation objectives. Although our fold-local preprocessing
reduces information leakage, the results remain conditional on a
single fitted synthetic generator and fixed model configurations.
Independent real-world validation and repeated generator and
anchor realizations are therefore necessary to establish broader
generalizability.

\section{Limitations and future work}
The evaluation relies on a single synthetic cohort generated from educational data with undocumented provenance, measurement windows, and sampling procedures. Future work should incorporate documented prospective datasets, independent generator replications, repeated anchor draws, and nested hyperparameter tuning.

\section{Conclusion}
Our leakage-aware comparison shows that learned-centroid
distances do not consistently outperform matched random anchors,
despite favorable descriptive clustering results. RF
achieves the strongest mean discrimination, while the usefulness
of augmentation depends on classifier, sample size and
evaluation metric. Attribution agreement and surrogate fidelity
also capture distinct properties. The study motivates matched
nonlinear controls when assessing cluster-derived predictors
and separate validation of descriptive segmentation, prediction
and explanation quality. Its conclusions remain conditional on
the fitted synthetic cohort.

\textbf{Declarations}\\
During this study, the authors used Generative AI (ChatGPT) to assist with experimental design, data generation, manuscript drafting, and language refinement.

\bibliographystyle{elsarticle-num}
\bibliography{bibliography}

\onecolumn
\begin{table*}[t]
\scriptsize
\centering\small\caption{Synthetic clustering sensitivity. Silhouette uses at most 2,000 observations; DBSCAN excludes noise from internal indices. A dash denotes an undefined index.}\label{tab:cluster-sensitivity}
\begin{tabular}{lrrrrrr}\toprule Algorithm & $k$ or $\epsilon$ & Min. samples & Clusters & Noise & Silhouette & DB\\\midrule
KMeans & 2 & 0 & 2 & 0 & 0.4297 & 0.9653\\
KMeans & 3 & 0 & 3 & 0 & 0.5034 & 0.6573\\
KMeans & 4 & 0 & 4 & 0 & 0.4965 & 0.6597\\
KMeans & 5 & 0 & 5 & 0 & 0.4289 & 0.7974\\
KMeans & 6 & 0 & 6 & 0 & 0.3959 & 0.9405\\
KMeans & 7 & 0 & 7 & 0 & 0.3574 & 0.9978\\
KMeans & 8 & 0 & 8 & 0 & 0.3461 & 1.0170\\
DBSCAN & 0.1 & 8 & 7 & 268 & -0.3573 & 0.7473\\
DBSCAN & 0.1 & 20 & 8 & 1483 & -0.0843 & 0.8110\\
DBSCAN & 0.1 & 50 & 6 & 4787 & 0.2161 & 0.6903\\
DBSCAN & 0.1 & 80 & 3 & 8030 & 0.4651 & 0.8221\\
DBSCAN & 0.15 & 8 & 2 & 79 & 0.1216 & 0.6878\\
DBSCAN & 0.15 & 20 & 1 & 294 & --- & ---\\
DBSCAN & 0.15 & 50 & 2 & 1689 & 0.3354 & 0.9081\\
DBSCAN & 0.15 & 80 & 4 & 3217 & 0.4820 & 0.7198\\
DBSCAN & 0.2 & 8 & 1 & 32 & --- & ---\\
DBSCAN & 0.2 & 20 & 1 & 96 & --- & ---\\
DBSCAN & 0.2 & 50 & 1 & 462 & --- & ---\\
DBSCAN & 0.2 & 80 & 2 & 1225 & 0.3321 & 0.9430\\
DBSCAN & 0.25 & 8 & 1 & 10 & --- & ---\\
DBSCAN & 0.25 & 20 & 1 & 39 & --- & ---\\
DBSCAN & 0.25 & 50 & 1 & 139 & --- & ---\\
DBSCAN & 0.25 & 80 & 1 & 400 & --- & ---\\
DBSCAN & 0.3 & 8 & 1 & 4 & --- & ---\\
DBSCAN & 0.3 & 20 & 1 & 15 & --- & ---\\
DBSCAN & 0.3 & 50 & 1 & 50 & --- & ---\\
DBSCAN & 0.3 & 80 & 1 & 124 & --- & ---\\
\bottomrule\end{tabular}
\end{table*}

\appendix

\section{Detailed predictive results}\label{app:pred}
\begin{center}\begin{minipage}{\textwidth}\centering\captionof{table}{Learning curves: mean AUROC across five folds for each training size.}

\begin{tabular}{lrrrrrrrr}\toprule & \multicolumn{2}{c}{LR} & \multicolumn{2}{c}{DT} & \multicolumn{2}{c}{RF} & \multicolumn{2}{c}{MLP}\\ Training size & Raw & Dist. & Raw & Dist. & Raw & Dist. & Raw & Dist.\\\midrule
500 & 0.8296 & 0.8278 & 0.7818 & 0.7843 & 0.8264 & 0.8266 & 0.7982 & 0.7779\\
1000 & 0.8325 & 0.8323 & 0.7992 & 0.7988 & 0.8370 & 0.8386 & 0.8163 & 0.8101\\
2000 & 0.8335 & 0.8343 & 0.8098 & 0.8177 & 0.8420 & 0.8416 & 0.8149 & 0.8183\\
4000 & 0.8346 & 0.8353 & 0.8152 & 0.8142 & 0.8488 & 0.8477 & 0.8265 & 0.8319\\
8000 & 0.8349 & 0.8359 & 0.8221 & 0.8227 & 0.8500 & 0.8528 & 0.8284 & 0.8363\\
\bottomrule\end{tabular}\end{minipage}\end{center}
\section{Explanation details}\label{app:xai}
\begin{center}\begin{minipage}{\textwidth}\centering\captionof{table}{Mean absolute finite-background SHAP and Mask-LIME coefficients on synthetic held-out cases (probability units).}\begin{tabular}{lrrrrrrrr}\toprule & \multicolumn{2}{c}{LR} & \multicolumn{2}{c}{DT} & \multicolumn{2}{c}{RF} & \multicolumn{2}{c}{MLP}\\ Feature & SHAP & Mask & SHAP & Mask & SHAP & Mask & SHAP & Mask\\\midrule
Age & 0.0167 & 0.0175 & 0.0128 & 0.0139 & 0.0125 & 0.0138 & 0.0127 & 0.0136\\
Gender & 0.0094 & 0.0097 & 0.0000 & 0.0000 & 0.0029 & 0.0031 & 0.0084 & 0.0090\\
Annual Income & 0.0335 & 0.0349 & 0.0218 & 0.0206 & 0.0304 & 0.0308 & 0.0329 & 0.0344\\
Spending Score & 0.0868 & 0.0900 & 0.1252 & 0.1243 & 0.1070 & 0.1077 & 0.0784 & 0.0796\\
Website Visits & 0.0428 & 0.0448 & 0.0394 & 0.0389 & 0.0445 & 0.0445 & 0.0341 & 0.0363\\
Time On Website & 0.0529 & 0.0550 & 0.0460 & 0.0483 & 0.0473 & 0.0487 & 0.0504 & 0.0520\\
Previous Purchases & 0.1115 & 0.1152 & 0.1089 & 0.1109 & 0.0896 & 0.0889 & 0.0848 & 0.0855\\
Satisfaction Score & 0.0130 & 0.0135 & 0.0101 & 0.0098 & 0.0181 & 0.0187 & 0.0205 & 0.0205\\
\bottomrule\end{tabular}\end{minipage}\end{center}
\section{Reproduction protocol}
The supplied synthetic CSV can be evaluated directly. The generation script additionally supports reconstruction from the private source file when the appropriate permissions are available. The generator uses seed 2026 and class-specific mixture seeds 2026 and 2027. Stratified folds use seed 42; nested subset and background sampling use seed $300+i$ for fold $i$. Four centroids use ten initializations and seed 42. The explanation audit uses 32 background rows, five test cases per fold, 256 masks and ridge penalty 0.01. All reported outputs derive from the single generated cohort. The archived data files contain the complete prediction scores, fold metrics, clustering sensitivity and attribution diagnostics.

\end{document}